\documentclass[lettersize,journal]{IEEEtran}
\usepackage{amsmath,amsfonts}
\usepackage{algorithmic}
\usepackage{array}
\usepackage[caption=false,font=normalsize,labelfont=sf,textfont=sf]{subfig}
\usepackage{textcomp}
\usepackage{stfloats}
\usepackage{url}
\usepackage{verbatim}
\usepackage{graphicx}
\def\BibTeX{{\rm B\kern-.05em{\sc i\kern-.025em b}\kern-.08em
    T\kern-.1667em\lower.7ex\hbox{E}\kern-.125emX}}
\usepackage{balance}

\usepackage{amsmath,amssymb,amsfonts}
\usepackage{algorithmic}
\usepackage{graphicx}
\usepackage{textcomp}
\usepackage{xcolor}

\usepackage{verbatim}
 \usepackage{color}
\usepackage{array}
\usepackage{longtable}
\usepackage{tablefootnote}
\usepackage{tikz}
\usepackage{amsmath,amssymb,amsfonts}
\usepackage{algorithmic}
\usepackage{setspace}  % For setting line spacing
\usepackage{lipsum}  % For generating dummy text
\usepackage{xcolor}

\usepackage[style=numeric,maxbibnames=1,sorting=none]{biblatex}
\usepackage{setspace}

\usepackage [ T1 ] { fontenc }
\usepackage [ linguistics ] {forest}

\usepackage[caption=false]{subfig}

\usepackage{lipsum}

\usepackage{titlesec}
   
\begin{document}
\titlespacing{\title}{0pt}{\parskip}{-\parskip}
%citations~\cite{IEEEexample:article_typical,IEEEexample:articleetal,IEEEexample:inproceedings_typical,IEEEexample:inproceedingsetal,IEEEexample:book_typical,IEEEexample:booketal,IEEEexample:misc_typical,IEEEexample:masterstype}

%\title{{\Large \textbf{Adaptive Continuous/Cumulative Adversarial Training (A\textunderscore CAT) for Machine/Deep Learning-Based SPAM Filters}}\\

%\title{{\Large \textbf{Enhancing The Robustness of SPAM Filters with ACAT: Adaptive Continuous Adversarial Training for Machine and Deep Learning}}\\

\title{{Introducing Adaptive Continuous Adversarial Training (ACAT) to Enhance ML Robustness\\{\small Authors’ draft for soliciting feedback: May 29, 2024}}\\

%\\{\small Authors’ draft for soliciting feedback: March 15, 2024}
%\\{\small Authors’ draft for soliciting feedback: March 15, 2024}

%Adaptive Continuous Adversarial Training (ACAT): Increasing ML and DL Robustness}}\\

}

\author{Mohamed elShehaby, Carleton University, Ottawa, Canada, mohamedelshehaby@cmail.carleton.ca\\Aditya Kotha, Carleton University, Ottawa, Canada, adityakotha@cunet.carleton.ca\\Ashraf Matrawy, Carleton University, Ottawa, Canada,  ashraf.matrawy@carleton.ca}

\begin{comment}

\author{\IEEEauthorblockN{Mohamed elShehaby}
\IEEEauthorblockA{
%\textit{dept. name of organization (of Aff.)} \\
\textit{Carleton University}\\
mohamedelshehaby@cmail.carleton.ca}
\and
\IEEEauthorblockN{Aditya Kotha}
\IEEEauthorblockA{
%{\textit{dept. name of organization (of Aff.)} \\
\textit{Carleton University}\\
adityakotha@cunet.carleton.ca}
\and
\IEEEauthorblockN{Ashraf Matrawy}
\IEEEauthorblockA{
%\textit{dept. name of organization (of Aff.)} \\
\textit{Carleton University}\\
ashraf.matrawy@carleton.ca}
}
\end{comment}
\maketitle

%\begingroup\renewcommand\thefootnote{\textsection}
%\footnotetext{This research was done while the author was doing a MITACS Globalink Research Internship at Carleton University.}
%\endgroup

\begin{abstract}

Adversarial training enhances the robustness of Machine Learning (ML) models against adversarial attacks. However, obtaining labeled training and adversarial training data in network/cybersecurity domains is challenging and costly. Therefore, this letter introduces Adaptive Continuous Adversarial Training (ACAT), a method that integrates adversarial training samples into the model during continuous learning sessions using real-world detected adversarial data. Experimental results with a SPAM detection dataset demonstrate that ACAT reduces the time required for adversarial sample detection compared to traditional processes. Moreover, the accuracy of the under-attack ML-based SPAM filter increased from 69\% to over 88\% after just three retraining sessions.

%Machine Learning (ML) is susceptible to adversarial attacks that aim to trick ML models, making them produce faulty predictions. Adversarial training was found to increase the robustness of ML models against these attacks. However, in network and cybersecurity, obtaining labeled training and adversarial training data is challenging and costly. Furthermore, concept drift deepens the challenge, particularly in dynamic domains like network and cybersecurity, and requires various models to conduct periodic retraining. This letter introduces Adaptive Continuous Adversarial Training (ACAT) to continuously integrate adversarial training samples into the model during ongoing learning sessions, using real-world detected adversarial data, to enhance model resilience against evolving adversarial threats. ACAT is an adaptive defense mechanism that utilizes periodic retraining to effectively counter adversarial attacks while mitigating catastrophic forgetting. Our approach also reduces the total time required for adversarial sample detection, especially in environments such as network security where the rate of attacks could be very high. Traditional detection processes that involve two stages may result in lengthy procedures. Experimental results using a SPAM detection dataset demonstrate that with ACAT, the accuracy of the SPAM filter increased from 69\% to over 88\% after just three retraining sessions. Furthermore, ACAT outperforms conventional adversarial sample detectors, providing faster decision times, up to four times faster in some cases. 
\end{abstract}


\section{Introduction}
\label{Intro}
%\textcolor{red}{what are teh main issues in current adverarial training. FOcus on those issues that ACAT addresses, do not list every single issue. }\\

%In  network and cybersecurity, acquiring ground truth labels for machine learning training datasets presents a challenging and costly endeavor \cite{apruzzese2022sok}. This scarcity of labeled training data in these domains exacerbates the shortage of adversarial training samples. 

%

The most popular machine learning (ML) approach is supervised learning  \cite{sarker2021machine}, where labeled datasets are used to train algorithms to classify data or predict outcomes accurately. However, in network and cybersecurity, acquiring ground truth (labels) for ML training datasets presents a challenging and costly endeavor \cite{apruzzese2022sok} \cite{ahmed2022machine}. ML is also prone to adversarial attacks that aim to trick ML models, making them produce faulty predictions. These attacks pose a significant threat to ML systems. As a defense against adversarial attacks, adversarial training, a process that involves augmenting training data with correctly labeled adversarial examples, has been shown to improve the robustness of ML models \cite{abou2020evaluation} \cite{abou2020investigating}. In adversarial training, training data is expanded by adding adversarial perturbations to some samples with known ground truth. The effectiveness of adversarial training is influenced by the quality and quantity of the original labeled data. Having more data allows for the generation of a wider variety of adversarial examples, which can potentially improve the model's robustness against such attacks. This is particularly relevant in network and cybersecurity domains, where scarcity of labeled training data can hinder the adversarial training process.

Another concern about the use of ML is ``concept drift'', which pertains to the unanticipated statistical changes in the characteristics of the prediction target \cite{lu2018learning}. Concept drift signifies that what is considered ``normal'' evolves over time, causing ML models to produce inaccurate predictions. In domains such as network and cybersecurity, where adversaries frequently modify their tactics \cite{casas2019should}, concept drift is highly prone to manifest.
To counter concept drift in ML, several techniques can be employed. One of the simplest and most intuitive methods is to periodically retrain the model using the most recent data. This ensures that the model adapts to the changes in the underlying data distribution.

\begin{figure}
\centering
\includegraphics[width=0.9\linewidth,keepaspectratio=true] {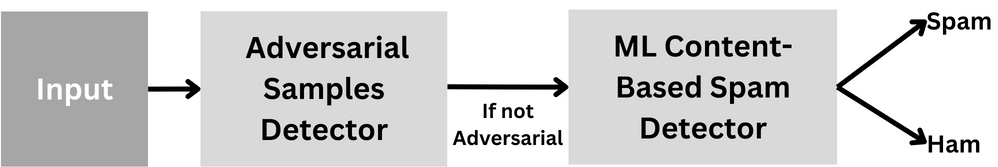}
	\caption{Conventional Detecting Approach}
	\label{fig:Para}
	\centering
\end{figure}

\begin{figure}
\centering
\includegraphics[width=1\linewidth,keepaspectratio=true] {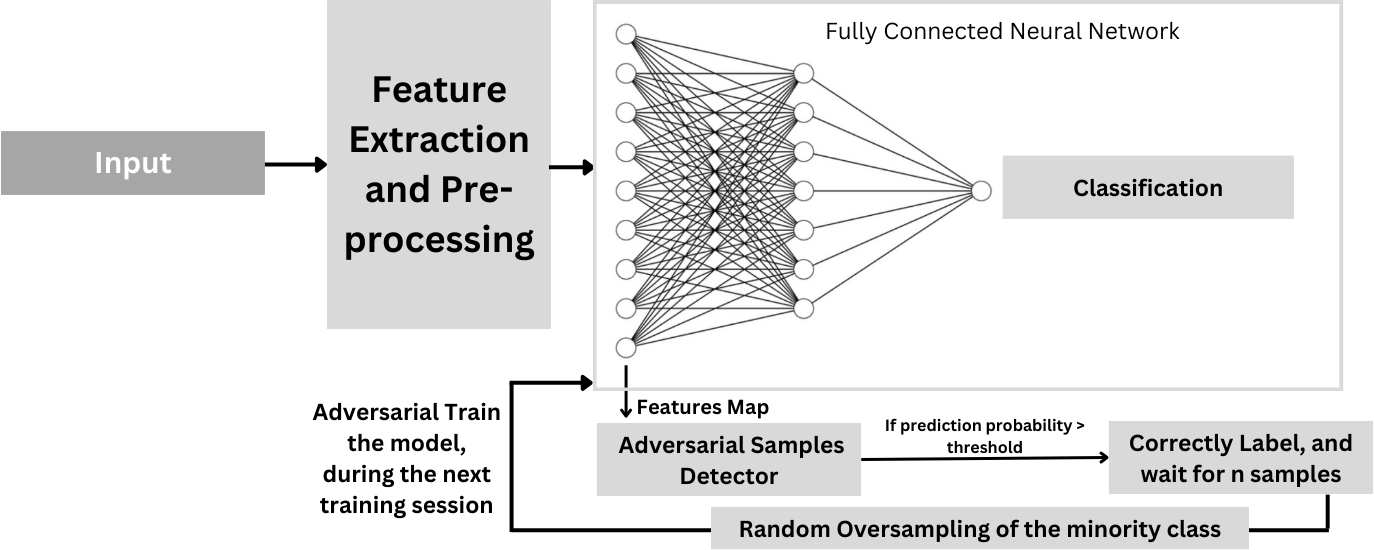}
\caption{The Proposed Adaptive Continuous Adversarial Training (ACAT)}
\label{fig:A_CAT}
\centering
\end{figure}

Our objective in this work is to continuously apply adversarial training samples to ML models during continuous training sessions, using real-world detected adversarial samples to increase the robustness of the models against evolving adversarial attacks. We present Adaptive Continuous Adversarial Training (ACAT) and make the following \textbf{contributions: }
\begin{itemize}
    \item Acts as an adaptive defence that uses continuous training.% It works against changing attacks. 
   % \item \textcolor{purple}{Addresses the problem of the lack of data for adversarial training because it uses attack data for training. It does not create new (e.g. artificial) attack data. }
    \item Addresses the problem of the lack of labeled data by using detected real-world attack data for adversarial training. This allows for a wider range of adversarial examples to be used for adversarial training, potentially improving robustness.
    \item Reduces the total time of adversarial sample detection, especially in environments such as network security where the rate of attacks could be very high. This is a significant improvement over conventional detection approache (Figure ~\ref{fig:Para}) that involve two stages and may result in lengthy procedures.  
    \item Deals with catastrophic forgetting \cite{kirkpatrick2017overcoming} during periodic continuous training
\end{itemize}

In order to evaluate ACAT, we used domain of SPAM filtering which required the following \textbf{contributions} that are specific to the experimental evaluation:  
\begin{itemize}
    \item Adapting the adversarial detection approach by Ye et al. \cite{ye2022feature} for the text-based SPAM problem.   
    \item Training the adversarial detector using a balanced dataset with an almost equal distribution of normal and adversarial samples, as well as between ham and SPAM samples.
\end{itemize}

\section{Related Work}

\subsection{Adversarial Evasion Attacks and Adversarial Training}

Evasion adversarial attacks, the primary focus of this letter, involve intentionally subtly modifying inputs to ML models in a way that causes them to misclassify or produce incorrect outputs. By analyzing the model's gradients, attackers can make targeted changes to the input that push the model's decision toward a wrong classification. In contrast, adversarial training is a defensive strategy used to enhance the robustness of ML models against adversarial attacks by incorporating adversarial examples into the training dataset, exposing the model to these deceptive inputs during the training process \cite{abou2020evaluation} \cite{abou2020investigating}.

\subsection{Continuous Machine Learning}

%As mentioned in Section \ref{Intro}, in machine learning, concept drift represents the unforeseen evolution of the statistical properties associated with the target variable. Put simply, ML models can get rusty over time. What was once ``normal'' can quickly shift, leading to inaccurate predictions. Concept Drift is especially risky and aggressive in dynamic domains like cybersecurity, where attackers constantly adapt their tactics. Consequently, 

To address concept drift, ML models must demonstrate adaptability, indicating that they need to continuously adjust. Several terms are employed to delineate these dynamic models, encompassing Online ML, Evolving ML, and Continuous ML.
There are diverse techniques for retraining, such as adaptive windowing \cite{bifet2007learning}, or simple ad-hoc continuous or periodic retraining. However, in dynamic learning environments, there is a risk of {\it catastrophic forgetting}  which is the phenomenon where a model, trained on a new task, significantly erases its previously acquired knowledge or performance on earlier tasks \cite{kirkpatrick2017overcoming}.

\section{Our Proposed Adaptive Continuous Adversarial Training (ACAT)}

%Concept drift is a foregone conclusion in ML applications, necessitating the implementation of continuous/dynamic ML techniques to mitigate its impact. Therefore, 

Our objective is to continuously apply adversarial training to the model during ongoing training sessions. In other words, our approach aims to analyze data while undergoing continuous adversarial training using an ML model designed to identify adversarial input samples, reducing the impact of adversarial attacks on ML models. As depicted in Figure \ref{fig:A_CAT}, the architecture of our proposed approach includes the following components:

%\subsection{Overview of pre-processing}

\noindent {\bf Pre-processing:} Pre-processing is the first step in ACAT. It is a crucial phase in ML involving the refinement of raw data to optimize its suitability for learning algorithms, thereby enhancing overall model performance. There are standard steps for pre-processing and in ACAT, the exact nature of pre-processing depends on the target applications domain (e.g. SPAM, network traffic, etc.). We explain our pre-processing for our SPAM problem evaluation in Section ~\ref{detailedPP}. 

%The steps of pre-processing depend on multiple factors such as the used ML model, application, training dataset, etc. These steps might include data cleaning, data transformation, data reduction, data balancing, feature scaling, tokenization, and stemming (text preprocessing techniques). Overall, effective preprocessing is pivotal for preparing diverse datasets for optimal ML model training and performance.

%\subsection{Do we need high level overview of LSTM or a generalization of it} \textcolor{red}{(Don't think so)}

%\noindent {\bf ML Model:} After proper data pre-processing, a suitable ML algorithm is selected based on the nature of the problem and the characteristics of the data. This could range from simple linear models to complex neural networks, depending on the required sophistication and the computational resources available. The next stage involves training the model, where the algorithm learns to make predictions or decisions based on the input data. This process is often iterative, requiring numerous adjustments to the model parameters to optimize performance. Once trained, the model is then rigorously tested and evaluated to assess its accuracy and effectiveness. The final step involves deploying the model into a real-world environment, where it can provide actionable insights or make decisions based on new data.

{\noindent {\bf ML Model:} After proper data pre-processing, a suitable ML algorithm is selected based on the nature of the problem and the characteristics of the data. In our ACAT, we primarily utilize neural networks because our continuous training uses the Elastic Weight Consolidation (EWC) method, which we discuss later in this section, requires models with weights.} The next stage involves training the model, where the Neural network learns to make predictions or decisions based on the input data. This process involves iterative fine-tuning, followed by testing for accuracy and effectiveness. Finally, the model is deployed for real-world applications.

%For time-series or sequential data, Long Short-Term Memory (LSTM) networks, a type of Recurrent Neural Network (RNN), are often employed due to their ability to capture temporal dependencies \cite{hochreiter1997long}.

%\subsection{Overview of adversarial detection}
%\textcolor{red}{In general, when the perturbations added to a sample are random, ML models tend to be robust to them.}

\noindent {\bf Adversarial detection:} Detecting samples with perturbations specifically crafted for an ML model is difficult \cite{feinman2017detecting}. This challenge is further amplified as these perturbations become more subtle. To address this issue, Ye et al. \cite{ye2022feature} proposed using feature maps from the layers preceding the Softmax layer in image models to improve the detection of changes in adversarial samples. Even if the perturbations are minor at the beginning, the feature maps from the original and perturbed samples often exhibit significant differences. Consequently, this facilitates the detection of adversarial samples. To ensure a balanced distribution of spam and ham labels of the detected adversarial samples, we performed random oversampling of the minority class with replacement. Subsequently, these balanced samples are used for the continuous adversarial training of the model during retraining sessions.

%\noindent {\bf Continuous training/Catastrophic forgetting:} Our Adaptive Continuous Adversarial Training (ACAT) approach implements Elastic Weight Consolidation (EWC) \cite{kirkpatrick2017overcoming} as a strategic approach to mitigate catastrophic forgetting. EWC offers a solution to this by effectively balancing the incorporation of new data with the retention of previously learned information. The core principle of EWC lies in identifying and preserving the importance of each parameter within the neural network. This is accomplished by calculating Fisher Information matrices, which quantify the significance of each model parameter in relation to the tasks it has previously learned. These matrices are used to generate a penalty term, which is integrated into the model’s loss function. By applying this technique, our approach can be sequentially adapted to new datasets while retaining cumulative knowledge from previous datasets. An alternative to EWC is fine-tuning, where we train our pre-trained model using a subset of adversarial data and manually adjust our model's hyperparameters in every retraining session. 

\noindent {\bf Continuous training/Catastrophic forgetting:} Our Adaptive Continuous Adversarial Training (ACAT) approach utilizes EWC \cite{kirkpatrick2017overcoming} as a strategic method to mitigate catastrophic forgetting. EWC effectively balances the integration of new data with the retention of previously learned information. This is achieved by calculating Fisher Information matrices from training data and new data batches which the model has previously adapted to. These matrices generate a penalty term that is integrated into the model’s loss function, enabling our approach to sequentially \textbf{adapt to new datasets while retaining knowledge from earlier ones.} An alternative to EWC is fine-tuning, where we train our pre-trained model using subsets of adversarial data and manually adjust hyperparameters during each retraining session. In both EWC and fine-tuning, we do not revert to using the entire dataset for retraining, which avoids excessive computational costs.
%%%%%%%%%%%%%%%%%%%%%%%%%%%%%%%%%%%%%%%%%%%%%%%%%%%%%%%%%%%%%%%%%%%%%%%%%%%%%%%%%%%%%%%%%%%%%%%%%%%%%%%%%%%%%%
\section{Experimental Evaluation using Problem-Space Adversarial Samples}

\subsection{Related work on ML in SPAM Filters}

SPAM filters are software or algorithms that detect and block unsolicited emails, commonly known as SPAM, from reaching a user’s inbox. They are an essential component of email services and clients, as they help users manage their emails more efficiently and protect them from phishing attempts, scams, and other malicious content. There are several techniques to detect SPAMs, including rule-based filters, content-based filters, header-based filters, and blacklist filters \cite{ahmed2022machine} \cite{dada2019machine}. Currently, ML significantly contributes to SPAM filtering, particularly within content-based filters \cite{dada2019machine}. Machine and deep learning algorithms enable experiential learning, resulting in the automatic generation of dynamic classification rules, which enhances the overall effectiveness of SPAM filtering. These adaptive and automated methods outperform traditional approaches such as blacklisting or rule-based filtering, which depend on manually crafted rules and are susceptible to evolving SPAM and phishing email tactics \cite{gangavarapu2020applicability}.

%Bring anything that is 
\subsection{Problem-space adversarial attacks against SPAM Filters}

Recent research studies have cast doubt on the practicality of adversarial attacks in the context of some applications in network and cybersecurity \cite{shehaby2023adversarial}\cite{el2023impact}. Within network security, there exist two adversarial attack spaces \cite{ibitoye2019threat}; In Feature-space attacks, the attacker modifies the input feature vector of the ML model by introducing perturbations to construct the attack. Conversely, in Problem-space attacks, the attacker introduces perturbations directly into the actual file itself, such as a network packet, to deceive the targeted model. Problem-space attacks are considered more realistic because gaining access to the model's feature vector input is typically implausible. Yet, perturbing the actual file, for instance, a network packet, can potentially disrupt network functionality or the intended behavior of the perturbed files. 

However, attacking ML-based SPAM detection systems might be less complicated. Well-crafted adversarial attacks involve subtly rephrasing a SPAM email using synonymous words and \textbf{introducing perturbations in the problem space.}
These alterations do not affect the malicious or network functionality of the sample; they primarily involve word substitutions. Researchers have explored practical adversarial attacks against SPAM filters. Gregory et al. \cite{gregory2023adversarial} proposed AG-WEP, a framework that dynamically selects perturbations to minimize modifications while misleading CNN-based classifiers with meaningful adversarial examples. Cheng et al. \cite{cheng2022adversarial} investigated vulnerabilities using adversarial examples, translating these perturbations into ``magic words'' \cite{wang2021crafting} to cause misclassification. These studies show the effectiveness of adversarial attacks against SPAM filters across models, datasets, and methods.

\subsection{Dataset and Adversarial Samples in our experiments}

Datasets such as Ling-SPAM, SpamAssassin, Enron-SPAM, and TREC07 are widely used in the Email domain to train SPAM filters \cite{janez2023review}. Ling-SPAM and SpamAssassin corpora have ham emails that may not be representative of typical user inboxes, while Enron-SPAM and TREC07 corpora offer better options for personalized SPAM filter development. We used the widely-adopted Enron-SPAM Corpus for training and testing due to its continued prevalence in spam email classification research \cite{guo2023spam}.

To create a practical problem-space black-box evasion adversarial attack (generated solely through model queries without any knowledge of its weights) and the dataset for training and evaluating the adversarial samples detector, we used the TextAttack library \cite{morris2020textattack} to adopt the TextFooler Attack by Jin et al. \cite{jin2020bert}. 
 %This implementation utilized WordSwapEmbedding for up to 50 replacements and constraints like RepeatModification, StopwordModification, and a WordEmbeddingDistance of 0.9. The attack executes via a GreedyWordSwapWIR search method, prioritizing impactful word replacement. 
Adversarial samples created by the TextFooler attack will be semantically similar to the original samples while bypassing the model. This effect is achieved by semantically replacing words with their counterparts, as this is a paraphrasing attack. 

%The attack functions by querying inputs and obtaining the output probabilities from the model, thereby making this attack very practical.

%To create a practical problem-space black-box evasion adversarial attack and the dataset for training and evaluating the adversarial samples detector, we used the TextAttack \cite{morris2020textattack} library to implement the TextFooler Attack by Jin et al. \cite{jin2020bert}.

As for the dataset and evaluation of the ML-based adversarial attacks detector, we used a mix of the adversarial samples and the SPAM dataset to create training data of nearly 1264 samples, maintaining an equal distribution between normal and adversarial, as well as ham and SPAM samples to avoid bias. The dataset pairs normal samples with their respective perturbed versions for training on our neural network model.

%\subsection{The Adversarial Attacks in Our Experiments}\label{adv-attack}
%less technical, and more info like black box problem space attack

 %This attack involves several key components, including a goal function for untargeted classification, constraints like RepeatModification, and WordEmbeddingDistance with a minimum cosine similarity of 0.9. %Additionally, we used the WordSwapEmbedding transformation and the GreedyWordSwapWIR search method, resulting in an effective adversarial attack strategy as depicted in Figure \ref{fig:SPAMEmail}. The TextFooler attack preserves semantic content and grammatical structure.

\subsection{Detailed pre-processing for the SPAM problem}
\label{detailedPP}

\begin{figure}
\centering

\includegraphics[width=0.7\linewidth,keepaspectratio=true]{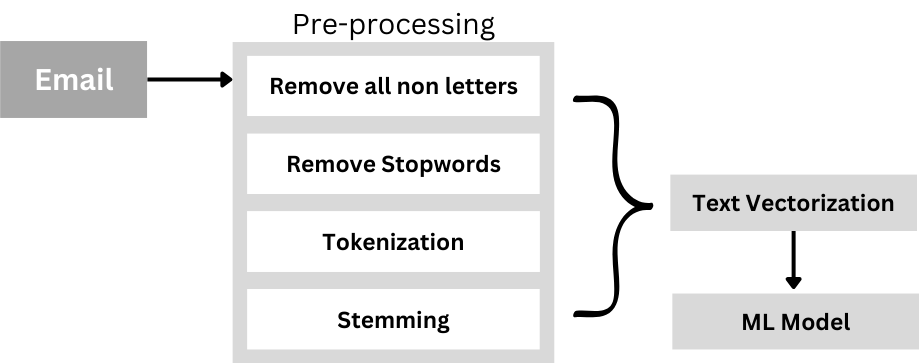}
\caption{Pre-processing of the Proposed approach, Adaptive Continuous Adversarial Training (ACAT)}
\label{fig:Pre-Pro}
\centering
\end{figure}

The pre-processing phase is crucial for text data preparation. It involves several key stages, as seen in Figure \ref{fig:Pre-Pro}. First, we remove non-letter characters, such as punctuation, symbols, and numbers, which simplifies the text and reduces noise. Next, we eliminate stop words, which are common, low-context words like ``and''  and ``the'' to enhance data quality and streamline downstream processes. After that, we tokenize the text, segmenting it into individual words or subword units, which facilitates structured representation and helps capture linguistic patterns. Finally, we apply stemming to reduce words to their base form, consolidating words with similar meanings, shrinking the vocabulary, and enhancing generalization. These pre-processing steps are crucial for ensuring that the text data is clean, structured, and ready for analysis. Then, a \textbf{Text Vectorization} phase is employed; In our current approach we used GloVe word embeddings \cite{pennington2014glove} with 100 dimension vectors to train a hybrid Bi-LSTM model. This choice leverages the strengths of RNNs in processing textual data, with LSTMs being particularly favored in spam research due to their effectiveness in handling sequences \cite{jain2019optimizing}. Training is performed on aggregated global word-word co-occurrence statistics from a corpus, and the resulting representations showcase interesting linear substructures of the word vector space. The advantage of GloVe is that, unlike Word2vec \cite{church2017word2vec}, GloVe does not rely just on local statistics (local context information of words), but incorporates global statistics (word co-occurrence) to obtain word vectors.

% GloVe is an unsupervised learning algorithm for obtaining vector representations for words.

%%%%%%%%%%%%%%%%%%%%%%%%%%%%%%%%%%%%%%%%%%%%%%%%%%%%%%%%%%%%%%%%%%%%%%%%%%%%%%%%%%%%%%%%%%%%%%%%%%%%%%%%%%%%%%
\section{Results and Discussion}

\subsection{Hybrid Bi-LSTM SPAM filter performance}

This section presents the results of testing the ML-based SPAM filter before applying adversarial attacks and continuous adversarial training. The trained hybrid Bi-LSTM model with word embeddings on the Enron SPAM corpus demonstrated an accuracy of 98.07\% on the test data, constituting 20\% of the 11,209 samples (these test samples were unperturbed). This accuracy was consistently observed across all evaluation metrics, including precision, recall, and F1 score, for both HAM and SPAM emails, all of which achieved a score of 0.98, as illustrated in Table \ref{tab:Enron-spam-performance}.

%The trained hybrid LSTM model with word embeddings on Enron SPAM corpus showed an accuracy of 98.07\% on the test data split from 20\% of 11,209 samples. This accuracy was consistently reflected across all evaluation metrics, including precision, recall, and F1 score, for both ham and SPAM emails, all of which achieved a score of 0.98, as shown in Table \ref{tab:Enron-spam-performance}.

\begin{table}
   \small	
  \centering
  \caption{Performance of the Hybrid Bi-LSTM SPAM Filter on Enron SPAM Test Corpus (Without Adversarial Samples)}
  \label{tab:Enron-spam-performance}
  \begin{tabular}{|c|c|c|c|c|}
    \hline
    \textbf{Class} & \textbf{Accuracy} & \textbf{Precision} & \textbf{Recall} & \textbf{F1-Score} \\
    \hline
    \textbf{Ham} & 0.98  & 0.98 & 0.98 & 0.98 \\
    \hline
    \textbf{SPAM} & 0.98  & 0.98 & 0.98 & 0.98 \\
    \hline
  \end{tabular}
\end{table}

\begin{table}
  \small	
  \centering
  \caption{Performance of Adversarial sample detector}
  \label{tab:adv-detecor-performance}
  \begin{tabular}{|c|c|c|c|c|}
    \hline
    \textbf{Class} & \textbf{Accuracy} & \textbf{Precision} & \textbf{Recall} & \textbf{F1-Score} \\
    \hline
    \textbf{Normal} & 0.93 &1.00 & 0.93 & 0.96 \\
    \hline
    \textbf{Adversarial}& 1.0 &0.93 & 1.00 & 0.96 \\
    \hline
  \end{tabular}
\end{table}

% as mentioned in Sub-Section \ref{adv-attack}

\subsection{Adversarial sample detector performance}

This section discusses the performance of the adversarial samples detector. Initially, we created TextFooler adversarial samples using the Enron SPAM dataset. We generated 505 adversarial samples to evade the Hybrid Bi-LSTM model, resulting in a total of 1010 samples consisting of normal and their respective perturbed samples. To ensure that we maintained the balanced distribution of SPAM and ham labels within our dataset, we employed oversampling of the minority class with replacement. This yielded an effective dataset of 1,264 samples for further analysis and model training. These samples were subsequently utilized to extract feature maps and assign labels indicating whether they were adversarial or not and are used to train the mentioned adversarial sample detector. This model showed an accuracy of 96.44\% on the 20\% test split of the 1264 samples. The complete performance of the adversarial sample detector is presented in Table \ref{tab:adv-detecor-performance}.

\subsection{ACAT vs Conventional Model}
\begin{figure}

\centering
\includegraphics[width=0.6\linewidth,keepaspectratio=true]{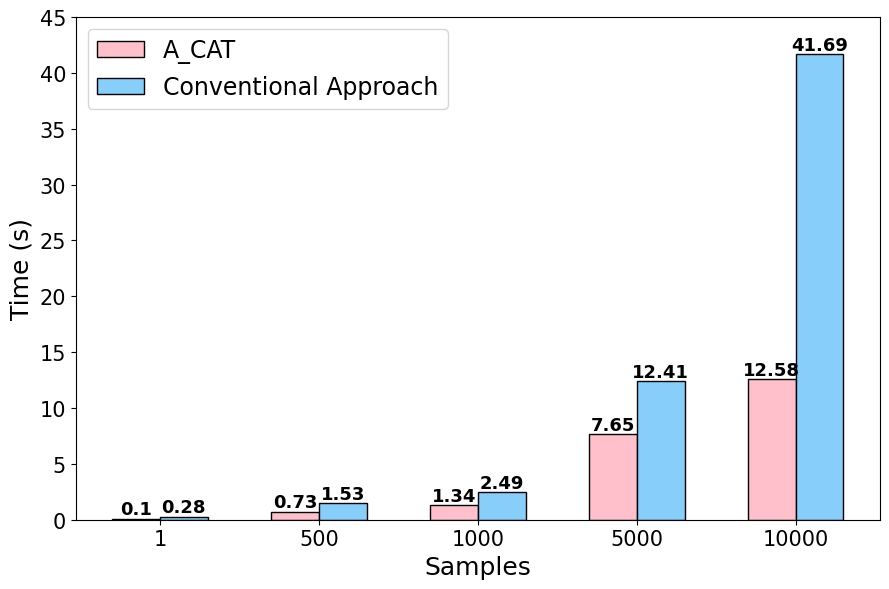}

	\caption{Prediction time of ACAT vs a Conventional Approach}
	\label{fig:time}
	\centering

\end{figure}

This section compares ACAT's prediction time to that of a conventional approach that is depicted in Figure \ref{fig:Para} which  utilizes the adversarial detector to predict final outputs instead of employing a continuous learning approach. However, as depicted in Figure \ref{fig:time}, which illustrates the prediction time of ACAT in comparison to the conventional approach, it becomes apparent that the classification time of the alternative method increases exponentially as the number of classified samples grows. For instance, with a single sample, the ACAT approach takes 0.1 seconds for classification, while the conventional approach requires approximately 0.28 seconds. Furthermore, as we scale up to 10000 samples, ACAT takes 12.58 seconds, whereas the conventional approach requires 41.69 seconds. This significant difference clearly underscores the efficiency of our approach. The discrepancy arises because capturing adversarial samples in ACAT occurs offline, while the alternative approach requires the use of feature maps during online inference, effectively doubling the inference time.

\begin{figure}

\centering
\includegraphics[width=0.6\linewidth,keepaspectratio=true]{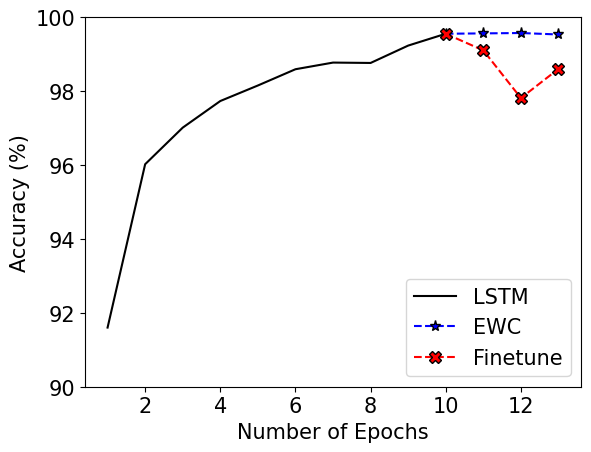}

	\caption{Accuracy of Fine Tuning vs EWC on the original training set (without adversarial perturbations) during Adversarial Continuous Training. The solid black line represents the 10 training epochs before deployment, while the dotted lines represent the accuracy after each adversarial training session.}
	\label{fig:acc}
	\centering

\end{figure}

\begin{figure}

\centering
\includegraphics[width=0.6\linewidth,keepaspectratio=true]{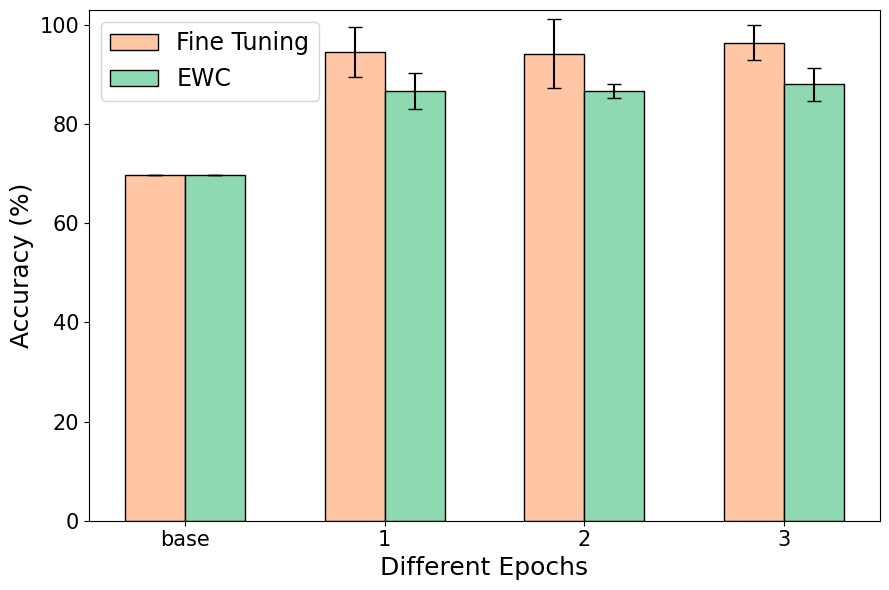}

	\caption{Effect of Continuous Adversarial Training: Accuracy Comparison on ($D_{\text{test}}$) of perturbed Enron SPAM (Fine Tuning vs EWC)}
	\label{fig:finetune}
	\centering

\end{figure}

\subsection{Fine Tuning vs EWC in Adversarial Continuous Training}

This section shows the effect of continuous adversarial training while actively countering the catastrophic forgetting. 
\noindent \textbf{Fine Tuning:} illustrates the accuracy of the models on the original, unperturbed training set during adversarial continuous training, testing for catastrophic forgetting. In the case of fine-tuning, a slight decrease in accuracy is evident. It dropped to as low as 97.81\% from its original accuracy of 99.55\%. It's worth noting that, in Figure \ref{fig:acc}, continuous training started after the ten training epochs of the LSTM model. Concurrently, Figure \ref{fig:finetune} illustrates the performance of our approach across three splits (three adversarial training sessions), each containing perturbed and unperturbed samples.  Before adversarial retraining, the model's accuracy on $D_{\text{test}}$ is 69.70\%. After fine-tuning for each epoch, the accuracy improves significantly: it reaches 94.544\% after the first epoch, then slightly decreases to 94.18\% in the second epoch, and increases again to 96.36\% by the third epoch. These results are the average of 5 different tests with data shuffling, and the standard deviations for these epochs are 1.795, 2.5, and 1.268, respectively. The confidence intervals for these accuracies, shown in Figure \ref{fig:finetune}, are calculated using the t-distribution, appropriate for the small sample size (n=5). 

\noindent \textbf{EWC:} In the case of the EWC, the accuracy remains almost constant on the Enron SPAM training data, as displayed in Figure \ref{fig:acc}, and also saw an increase in training accuracy to 99.57\% from the base of 99.55\% at the 2nd iteration. This stability is also observed in the continuous training of the hybrid Bi-LSTM model on the new adversarial sample data. In the first two iterations, the model's accuracy improves to 86.66\%, and further increases to 88.0\% by the third epoch, as depicted in Figure \ref{fig:finetune}. The performance across these epochs is derived as mentioned above, with the standard deviations recorded as 1.272, 0.541, and 1.176 across the three epochs, respectively. The confidence intervals are also shown in Figure \ref{fig:finetune}. \textbf{Summary: Unperturbed data:} EWC entirely avoids catastrophic forgetting and exhibits lower performance variability across epochs compared to fine-tuning (Figure \ref{fig:acc}). \textbf{Perturbed data:} Both methods improve robustness after retraining (Figure \ref{fig:finetune}). While the confidence intervals mostly overlap (indicating that the difference between them is insignificant), EWC has the advantage of an automated, hassle-free retraining process.

\section{Conclusion}

%In this letter, we introduce ACAT as a novel approach to enhance the resilience of ML systems against adversarial attacks. ACAT continuously integrates adversarial training samples into the model during ongoing learning sessions, using real-world detected adversarial data. Moreover, ACAT mitigates catastrophic forgetting during periodic continuous training. Our work highlights the importance of considering the dynamic nature of modern ML models to create a continuous adversarial training defense system capable of adapting to evolving attacks over time.
In this letter, we introduce ACAT, a novel approach to enhance the resilience of ML systems against adversarial attacks. ACAT continuously integrates real-world detected adversarial data into the model during ongoing learning sessions while mitigating catastrophic forgetting during periodic continuous training. Our work highlights the importance of considering the dynamic nature of modern ML models to create a continuous adversarial training defense system capable of adapting to evolving attacks over time. 

Using a SPAM detection dataset, our results show that our adversarial detector, which captures adversarial samples for continuous adversarial training and was trained using the dataset we created, achieved an F1-measure of 0.96. Additionally, after only three retraining sessions, the accuracy of our attacked SPAM filter increased from 69\% to 88\%. Our results indicate that ACAT gained cumulative and continuous knowledge about adversarial samples without catastrophic forgetting. Compared to the conventional approach of using two separate models for adversarial detection and SPAM detection, our proposed ACAT is faster at decision time (up to four times faster when handling 10,000 samples).

\section{ Acknowledgement}
{This research was done while the second author was doing an internship at Carleton University through the support of the MITACS Globalink Research Internship program. The First and third authors acknowledge support from the NSERC Discovery Grant program.}

 % \bibliographystyle{abbrv}
 %\bibliographystyle{ieeetr}
 
%\bibliography{mainICC}

\begingroup
\setstretch{0.8}
\setlength\bibitemsep{0pt}
\printbibliography
\endgroup

\end{document}